\documentclass[letterpaper,11pt,onecolumn]{article}
\usepackage[dvips]{graphicx}
\usepackage[dvips]{floatflt}

\newcommand{\be}{\begin{equation}}
\newcommand{\ee}{\end{equation}}
\newcommand{\beq}{\begin{equation}}
\newcommand{\eeq}{\end{equation}}
\newcommand{\bed}{\begin{displaymath}}
\newcommand{\eed}{\end{displaymath}}
\newcommand{\beqa}{\begin{eqnarray}}
\newcommand{\eeqa}{\end{eqnarray}}
\newcommand{\beqann}{\begin{eqnarray*}}
\newcommand{\eeqann}{\end{eqnarray*}}
\newcommand{\bseq}{\begin{subequation}}
\newcommand{\eseq}{\end{subequation}}

\newcommand{\ba}{\begin{array}}
\newcommand{\ea}{\end{array}}

\newcommand{\negr}[1]{{\bf {#1}}}

\makeatletter
\def\@normalsize{\@setsize\normalsize{12pt}\xpt\@xpt
\let\@listi\@listI}
\belowdisplayskip\abovedisplayskip
\def\subsize{\@setsize\subsize{12pt}\xipt\@xipt}
\def\section{\@startsection {section}{1}{\z@}{-3ex plus -1ex minus
 -.2ex}{1.0ex plus .2ex}
{\Large}}
\def\subsection{\@startsection {subsection}{2}{\z@}{-2ex plus
    -.5ex minus  -.2ex}{1.0ex plus .2ex}{\Large}}
\def\subsubsection{\@startsection {subsubsection}{3}{\parindent}{1.5ex plus
.2ex minus .2ex}{.5ex plus .2ex}{\normalsize\sl}}
\def\paragraph{\@startsection
 {paragraph}{4}{\z@}{3.25ex plus 1ex minus .2ex}{-1em}{\normalsize\bf}}
\def\subparagraph{\@startsection
 {subparagraph}{4}{\parindent}{3.25ex plus 1ex minus
 .2ex}{-1em}{\normalsize\bf}}
\makeatother
\font\tencmr=cmr10
\font\twelvecmr=cmr12

\def\figurename{\tencmr Fig.}
\begin{document}
\date{}
\title{{\Large\uppercase {Tenth World Congress On The Theory Of}} \\
       {\Large\uppercase {Machines And Mechanisms}} \\
       {\twelvecmr Oulu, Finland, June 20-24, 1999} \\
       {\Large \uppercase {On the characterization of the regions of
       feasible trajectories in the workspace of parallel
       manipulators}}}
\maketitle
\vspace{-0.5cm}
\begin{Auteur}
  \begin{center}
      \begin{tabular}[t]{cc}
          Damien Chablat$^1$ &  Philippe Wenger$^2$
      \end{tabular} \\
      \vspace{0.5cm}
      \begin{tabular}[t]{ccc}
          {\tencmr $^1$McGill Centre for Intelligent Machines} &
          {\tencmr $^1$INRIA Rocquencourt} &
          {\tencmr $^2$Institut de Recherche} \\
          {\tencmr McGill University} &
          {\tencmr Domaine de Voluceau} &
          {\tencmr en Cybern\'etique de Nantes} \\
          {\tencmr 817 Sherbrooke Street West} &
          {\tencmr B.P. 105} &
          {\tencmr 1, rue de la No\"e} \\
          {\tencmr Montreal, Quebec, Canada H3A 2K6} &
          {\tencmr 78 153 Le Chesnay France} &
          {\tencmr 44321 Nantes, France}
      \end{tabular} \\
      \begin{tabular}[t]{cc}
          {\tencmr Chablat@cim.mcgill.ca} &
          {\tencmr Philippe.Wenger@ircyn.ec-nantes.fr}
      \end{tabular}
  \end{center}
\end{Auteur}
\section{Introduction}
It was shown recently that parallel manipulators with several
inverse kinematic solutions have the ability to avoid parallel
singularities [Chablat 1998a] and self-collisions [Chablat 1998b]
by choosing appropriate joint configurations for the legs. In
effect, depending on the joint configurations of the legs, a given
configuration of the end-effector may or may not be free of
singularity and collision. Characterization of the
collision/singularity-free workspace is useful but may be
insufficient since two configurations can be accessible without
collisions nor singularities but it may not exist a feasible
trajectory between them.
\par
The goal of this paper is to define the maximal regions of the
workspace where it is possible to execute trajectories. Two
different families of regions are defined : 1. those regions where
the end-effector can move between any set of points, and 2. the
regions where any continuous path can be tracked. These regions are
characterized from the notion of aspects and free-aspects recently
defined for parallel manipulators [Chablat 1998b]. The construction
of these regions is achieved by enrichment techniques and using an
extension of the octree structures to spaces of dimension greater
than three. Illustrative examples show the interest of this study
to the optimization of trajectories and the design of parallel
manipulators.
\section{Preliminaries}
\subsection{Kinematics}
The input vector \negr q (the vector of actuated joint values) is
related to the output vector \negr X (the vector of configuration
of the moving platform) through the following general equation :
\be
        F(\negr X, \negr q)=0
        \protect\label{equation:the_kinematic}
\ee
Vector (\negr X, \negr q) will be called {\em manipulator
configuration} and \negr X is the platform configuration and will
be more simply termed {\em configuration}. Differentiating equation
(\ref{equation:the_kinematic}) with respect to time leads to the
velocity model
\be
     \negr A \negr t + \negr B \dot{\negr q} = 0
\ee
With $\negr t=\left[w, \dot{\negr c} \right]^T$, for planar
manipulators ($w$ is the scalar angular-velocity and $\dot{\negr
c}$ is the two-dimensional velocity vector of the operational point
of the moving platform), $\negr t=\left[\negr w \right]^T$, for
spherical manipulators and $\negr t=\left[\negr w,\dot{\negr
c}\right]^T$, for spatial manipulators ($\dot{\negr c}$ is the
three-dimensional velocity vector and $\negr w$ is the
three-dimensional angular velocity-vector of the operational point
of the moving platform).
\par
Moreover, \negr A and \negr B are respectively the
direct-kinematics and the inverse-kinematics matrices of the
manipulator. A singularity occurs whenever \negr A or
\negr B, (or both) can no longer be inverted. Three
types of singularities exist [Gosselin 1990]:
\beqa
    (1)\quad det(\negr A) = 0 \quad (2) \quad det(\negr B) = 0 \quad
    (3)\quad det(\negr A) = 0 \quad and \quad det(\negr B) = 0  \nonumber
\eeqa
\subsection{Parallel and serial singularities}
Parallel singularities occur when the determinant of the direct
kinematics matrix \negr A vanishes. The corresponding singular
configurations are located inside the workspace. They are
particularly undesirable because the manipulator can not resist any
effort and control is lost.
\par
Serial singularities occur when the determinant of the inverse
kinematics matrix \negr B vanishes. By definition, the
inverse-kinematic matrix is always diagonal: for a manipulator with
$n$ degrees of freedom, the inverse kinematic matrix \negr B can be
written as $\negr B = Diag\left[\negr B_{11}, ..., \negr
B_{jj}, ..., \negr B_{nn}\right]$, each term $\negr B_{jj}$ being
associated with one leg. A serial singularity occurs whenever at
least one of these terms vanishes. When the manipulator is in
a serial singularity, there is a direction along which no Cartesian
velocity can be produced.
\subsection{Point-to-point trajectory}
There are two major types of tasks to consider~: point-to-point
trajectories and continuous trajectories. We consider in this section
point-to-point trajectories.
\begin{Def}
A point-to-point trajectory $T_{pp}$ is defined by a set of $p$
configurations in the workspace~: $T_{pp} = \{\negr X_1,..., \negr
X_i,
....,  \negr X_p\}$.
\end{Def}
By definition, no path is prescribed between any two configurations
$\negr X_i$ and $\negr X_j$.
\begin{Hyp}
In a point-to-point trajectory, the moving platform can not move
through a parallel singularity.
\end{Hyp}
\par
Although it was shown recently that in some particular cases a
parallel singularity could be crossed [Nenchev 1997], hypothesis 1
is set for the most general cases.
\par
           A point-to-point trajectory $T_{pp}$ will be feasible if there
           exists a continuous path in the Cartesian product of the workspace
           by the joint space which does not meet a parallel singularity and
           which makes the moving platform pass through all prescribed
           configurations $\negr X_i$ of the trajectory $T_{pp}$.
\begin{figure}[hbt]
    \hspace{-0.3cm}
    \begin{tabular}{cc}
       \begin{minipage}{100 mm}
           \begin{Rem}
           A fully parallel manipulator with several inverse kinematic
           solutions can change its joint configuration between two prescribed
           configurations. Such a manoeuver may enable the manipulator to
           avoid a parallel singularity (Figure
           \ref{figure:singularity_regular_configurations}). More generally,
           the choice of the joint configuration for each configuration
           $\negr X_i$ of the
           trajectory $T_{pp}$ can be established by any other criteria like
           stiffness or cycle time [Chablat et al. 1998]. Note that a change of
           joint configuration makes the manipulator run into a serial
           singularity, which
           is not redhibitory for the feasibility of point-to-point
           trajectories.
           \end{Rem}
       \end{minipage} &
       \begin{minipage}{55 mm}
         \begin{center}
           \includegraphics[width= 38mm,height= 26.8mm]{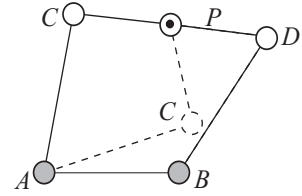}
           \caption{Singular (solid) and a regular (dotted) configurations
           (the actuated joints are $A$ and $B$)}
           \protect\label{figure:singularity_regular_configurations}
         \end{center}
       \end{minipage}
    \end{tabular}
\end{figure}
\subsection{Continuous Trajectory}
\begin{Def}
A continuous trajectory $T_{c}$ is defined by a parametric
curve in the operational space as~: $T_{c}=
\lambda([0, 1])$ where $\lambda$ is a continuous function defined on [0, 1]
and differentiable by parts on this interval.
\end{Def}
\begin{Hyp}
In a continuous trajectory, the moving platform cannot meet a parallel
singularity nor a serial singularity.
\end{Hyp}
\par
A continuous trajectory $T_{c}$ will be feasible if there
exists a continuous path in the Cartesian product of the workspace
by the joint space which is free of parallel and serial
singularities and which makes the platform move along the
continuous trajectory $T_{c}$.
\section{The N-connected regions}
\begin{Def}
The {\em N-connected regions} of the workspace are the maximal
regions where any point-to-point trajectory is feasible.
\end{Def}
For manipulators with multiple inverse and direct kinematic
solutions, it is not possible to study the joint space and the
workspace separately. First, we need to define the {\em regions of
manipulator reachable configurations} in the Cartesian product of
the workspace by the joint space $W.Q$.
\begin{Def}
The regions of manipulator reachable configurations $R_j$ are
defined as the maximal sets in $W . Q$ such that
\end{Def}
\begin{itemize}
\item $R_j \in W . Q$,
\vspace{-0.2cm}
\item $R_j$ is connected,
\vspace{-0.2cm}
\item $R_j= \{\negr X, \negr q\}$ such that $det(\negr A) \neq 0$
\end{itemize}
In other words, the regions $R_j$ are the sets of all
configurations (\negr X, \negr q) that the manipulator can reach
without meeting a parallel singularity and which can be linked by a
continuous path in $W . Q$.
\begin{Prop}
A trajectory $T_{pp}=\{\negr X_1,...,\negr X_p\}$ defined in the
workspace $W$ is feasible if and only if :
\end{Prop}
\beqa
\forall \negr X \in \{\negr X_1,..., \negr X_p\},
\quad \exists \negr q_i \in Q, \exists R_j
\quad such~that~(\negr X_i, \negr q_i) \in R_j \nonumber
\eeqa
In other words, for each configuration $\negr X_i$ in $T_{pp}$,
there exists at least one joint configuration $\negr q_i$ and one region of
manipulator reachable configurations $R_j$ such that the
manipulator configuration $(\negr X_i, \negr q)$ is in $R_j$.
\begin{Pro}
Indeed, if for all configurations  $\negr X_i$, there is one joint
configuration $\negr q_i$ such that $(\negr X_i, \negr q_i) \in
R_j$ then the trajectory is feasible because, by definition, a
region of manipulator reachable configurations is connected and
free of parallel singularity. Conversely, if for a given
configuration $\negr X_i$, it is not possible to find a joint configuration
$\negr q_i$ such that $(\negr X_i, \negr q_i) \in R_j$, then no
continuous, parallel singularity-free path exists in $W . Q$ which
can link the other prescribed configurations.
\end{Pro}
\vspace{0.5cm}
\begin{Theo}
The N-connected regions $W_{N j}$ are the projection ${\bf \Pi}_W$
of the regions of manipulator reachable configurations $R_j$ onto
the workspace:
\end{Theo}
\beqa
W_{N j}= {\bf \Pi}_W R_j \nonumber
\eeqa
\begin{Pro}
This results is a straightforward consequence of the above
proposition.
\end{Pro}
\par
The N-connected regions cannot be used directly for planning
trajectories in the workspace since it is necessary to choose one
joint configuration $\negr q$ for each configuration $\negr X$ of
the moving platform such that $(\negr X, \negr q)$ is included in
the same region of manipulator reachable configurations $R_j$.
However, the N-connected regions provide interesting global
information with regard to the performances of a fully parallel
manipulators because they define the maximal regions of the
workspace where it is possible to execute any point-to-point
trajectory.
\par
A consequence of the above theorem is that the workspace $W$ is
N-connected if and only if there exists a N-connected region $W_{N
j}$ which is coincident with the workspace :
\beqa
W_{N j} = W \nonumber
\eeqa
\section{The T-connected regions}
\begin{Def}
The {\em T-connected regions} of the workspace are the maximal
regions where any continuous trajectory is feasible.
\end{Def}
\par
When the mobile platform moves along a continuous trajectory, a
fully parallel manipulator cannot pass through a serial or a
parallel singularity. The aspects $\negr A_{ij}$ [Chablat 1998b]
characterize the maximal domains without serial and parallel
singularity.
\par
We recall below the definition of an aspect :
\begin{itemize}
\item $\negr A_{ij} \subset W \cdot Q$;
\vspace{-0.2cm}
\item $\negr A_{ij}$ is path-connected;
\vspace{-0.2cm}
\item $\negr A_{ij} = \left\{
                        (\negr X, \negr q) \subset W \cdot Q ~such~that~
                          det(\negr A) \neq 0 ~and~ det(\negr B) \neq 0
                      \right\}$
\end{itemize}
\begin{Prop}
A continuous trajectory $T_{c}=\{\negr X_\lambda, \lambda
\in [0, 1])\}$ defined in the workspace $W$ is feasible if and only if :
\end{Prop}
\beqa
\forall \lambda \in [0,1], \exists \negr q \in Q \quad such~that \quad
(\negr X_\lambda, \negr q) \in \negr A_{ij} \nonumber
\eeqa
In other words, for all configurations $\negr X_\lambda$ in
$T_{c}$, there exists at least one joint configuration $\negr q$
such that the manipulator configuration $(\negr X_\lambda, \negr
q)$ is in $\negr A_{ij}$.
\begin{Pro}
Indeed, if for all configurations  $\negr X_\lambda$, there is one
joint configuration $\negr q$ such that $(\negr X_\lambda, \negr q)
\in \negr A_{ij}$ then the trajectory is feasible because, by definition, an
aspect is connected and free of serial and parallel singularity.
Conversely, if for a given configuration $\negr X_\lambda$, it is
not possible to find a joint configuration $\negr q$ such that
$(\negr X_\lambda, \negr q) \in \negr A_{ij}$, then no continuous
singularity-free path exists in $W . Q$ which can link the other
prescribed configurations.
\end{Pro}
\vspace{0.5cm}
\begin{Theo}
The T-connected regions $W_{T j}$ are the projection ${\bf \Pi}_W$
of the aspect $\negr A_{ij}$ onto the workspace:
\end{Theo}
\beqa
W_{T j}= {\bf \Pi}_W \negr A_{ij} \nonumber
\eeqa
\begin{Pro}
This results is a straightforward consequence of the above
proposition.
\end{Pro}
A consequence of the above theorem is that the workspace $W$ is
T-connected if and only if there exists a T-connected region $W_{T
j}$ which is coincident with the workspace :
\beqa
W_{T j} = W \nonumber
\eeqa
\section{Example: A Two-DOF fully parallel manipulator}
For more legibility, a planar manipulator is used as illustrative
example in this paper. This is a five-bar, revolute
($R$)-closed-loop linkage, as displayed in
figure~\ref{figure:manipulateur_general}. The actuated joint
variables are $\theta_1$ and $\theta_2$, while the Output values
are the ($x$, $y$) coordinates of the revolute center $P$. The
passive joints will always be assumed unlimited in this study.
Lengths $L_0=7$, $L_1=8$, $L_2=5$, $L_3=8$, and $L_4=5$ define the
geometry of this manipulator entirely, in certain units of length
that we need not specify. The actuated joints are not unlimited
$\theta_1, \theta_2 \in [0,\pi]$. In order to point out that the
workspace of a parallel manipulator may not be N-connected even
when there is no obstacle obstructions, it is assumed here that the
manipulator is free of self-collisions. Examples with collision
analyses have been investigated and can be found in the PhD thesis
of the first author of this paper [Chablat 1998c].
\par
The workspace is shown in figure \ref{figure:workspace}. We want to
know whether this manipulator can execute any point-to-point motion
or continuous trajectory in the workspace. To answer this question,
we need to determine the the N-connected regions and the
T-connected regions.
\subsection{N-connected regions}
\begin{figure}[hbt]
    \begin{tabular}{ccc}
       \begin{minipage}{50 mm}
         \begin{center}
           \includegraphics[width= 45mm,height= 31.5mm]{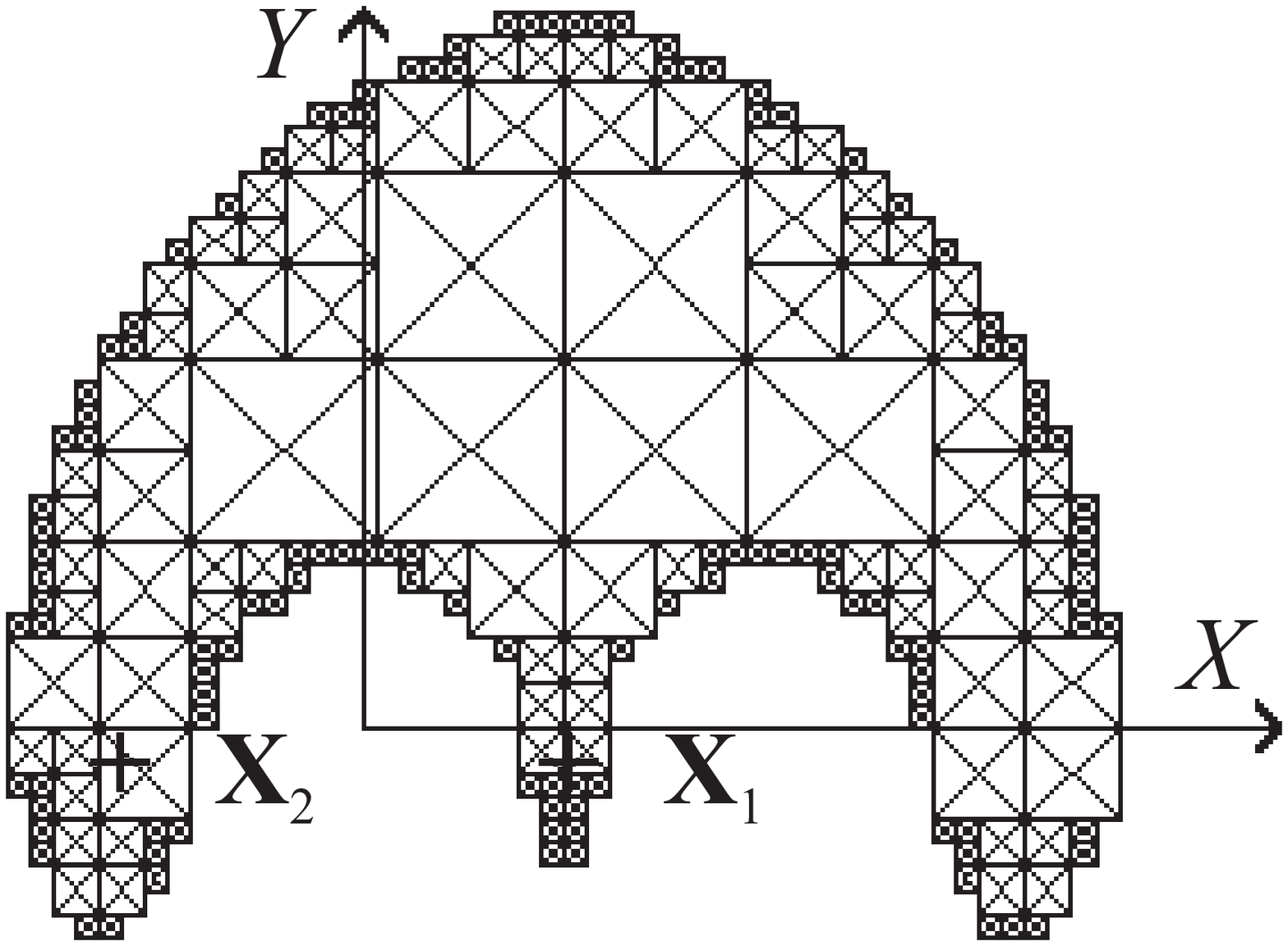}
           \caption{The workspace}
           \vspace{0.4cm}
           \protect\label{figure:workspace}
         \end{center}
       \end{minipage} &
       \begin{minipage}{50 mm}
         \begin{center}
           \includegraphics[width= 45mm,height= 28.8mm]{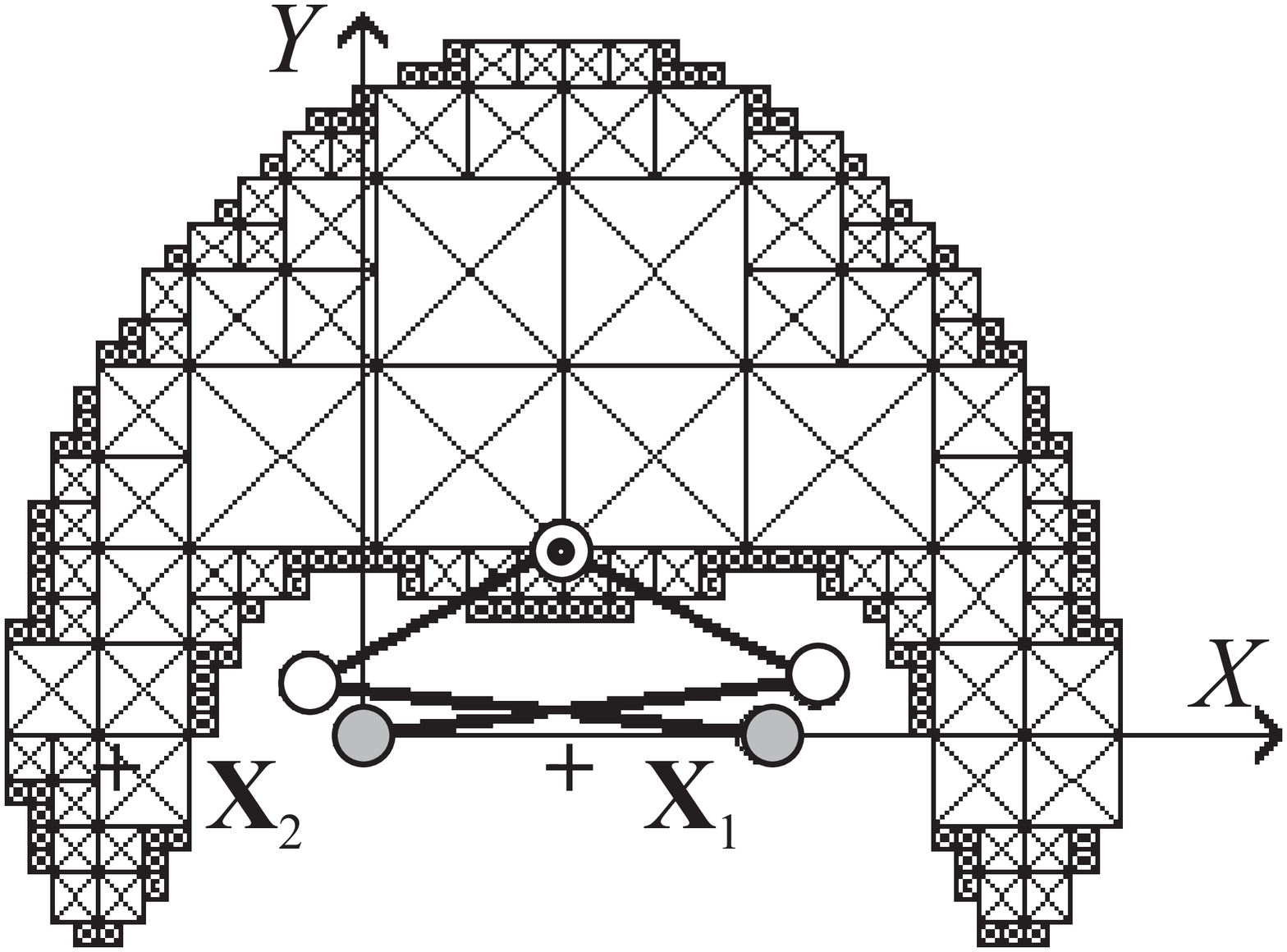}
           \caption{The first N-connected region of the workspace}
           \protect\label{figure:First_N_connected_region}
         \end{center}
       \end{minipage} &
       \begin{minipage}{50 mm}
         \begin{center}
           \includegraphics[width= 45mm,height= 28.8mm]{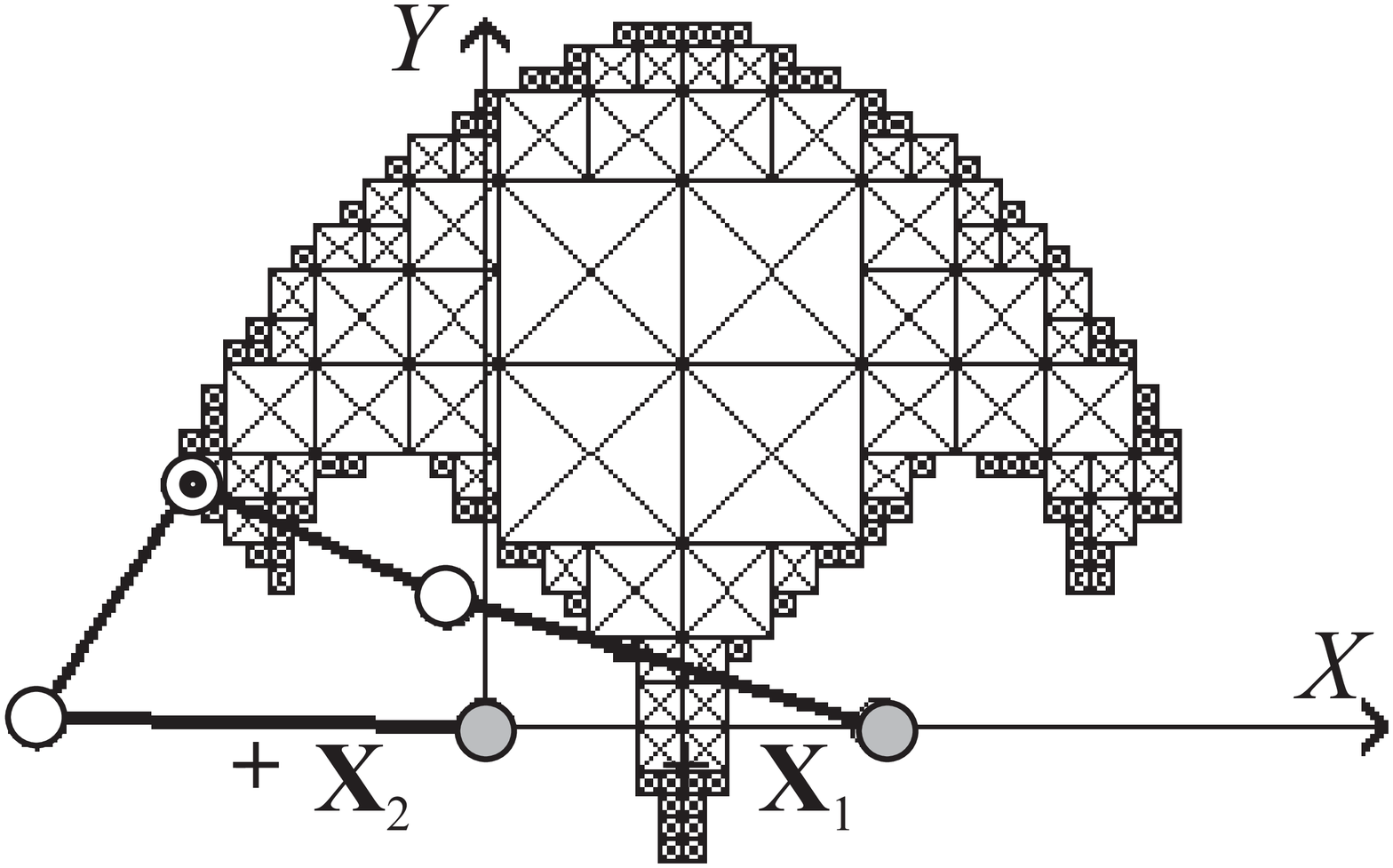}
           \caption{The second N-connected region of the workspace}
           \protect\label{figure:Second_N_connected_region}
         \end{center}
       \end{minipage}
    \end{tabular}
\end{figure}

\begin{floatingfigure}{55mm}
       \begin{center}
         \includegraphics[width= 50.1mm,height= 47.5mm]{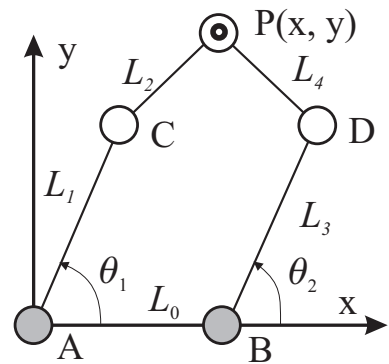}
         \caption{A two-dof fully parallel manipulator}
         \protect\label{figure:manipulateur_general}
       \end{center}
\end{floatingfigure}
It turns out, although there is no obstacles obstructions, that the
workspace of the manipulator at hand is not N-connected, e.g. the
manipulator cannot move its platform between any set of
configurations in the workspace. In effect, due to the existence of
limits on the actuated joints, not all joint configurations are
accessible for any configuration in the workspace. Thus, the
manipulator may lose its ability to avoid a parallel singularity
when moving from one configuration to another. This is what happens
between points $\negr X_1$ and $\negr X_2$ (Figure
\ref{figure:workspace}). These two points cannot be linked by the
manipulator although they lie in the workspace which is connected
in the mathematical sense (path-connected) but not N-connected. In
fact, there are two separate N-connected regions which do not
coincide with the workspace and the two points do not belong to the
same N-connected region (Figures
\ref{figure:First_N_connected_region} and
\ref{figure:Second_N_connected_region}). We know that the workspace
of serial manipulators is always N-connected when there is no
collision [Wenger 1991]. The example treated in this section shows
that this is false when the manipulator is parallel.
\vspace{-0.2cm}
\subsection{T-connected regions}
\begin{figure}[hbt]
    \begin{tabular}{cccc}
       \begin{minipage}{36 mm}
         \begin{center}
           \includegraphics[width= 36mm,height= 22.8mm]{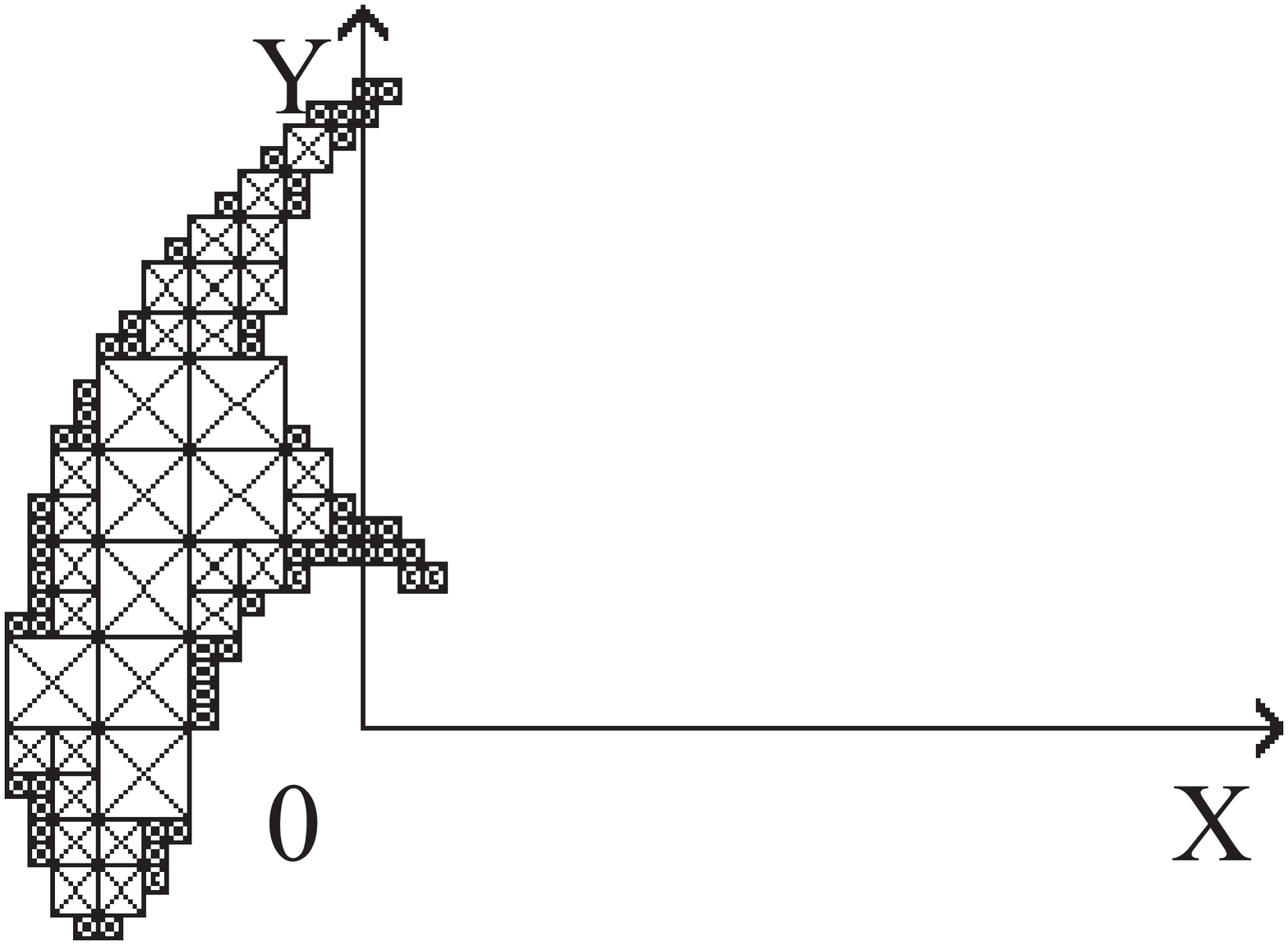}
           \caption{Joint configuration 1}
           \protect\label{figure:posture_1}
         \end{center}
       \end{minipage} &
       \begin{minipage}{36 mm}
         \begin{center}
           \includegraphics[width= 36mm,height= 22.8mm]{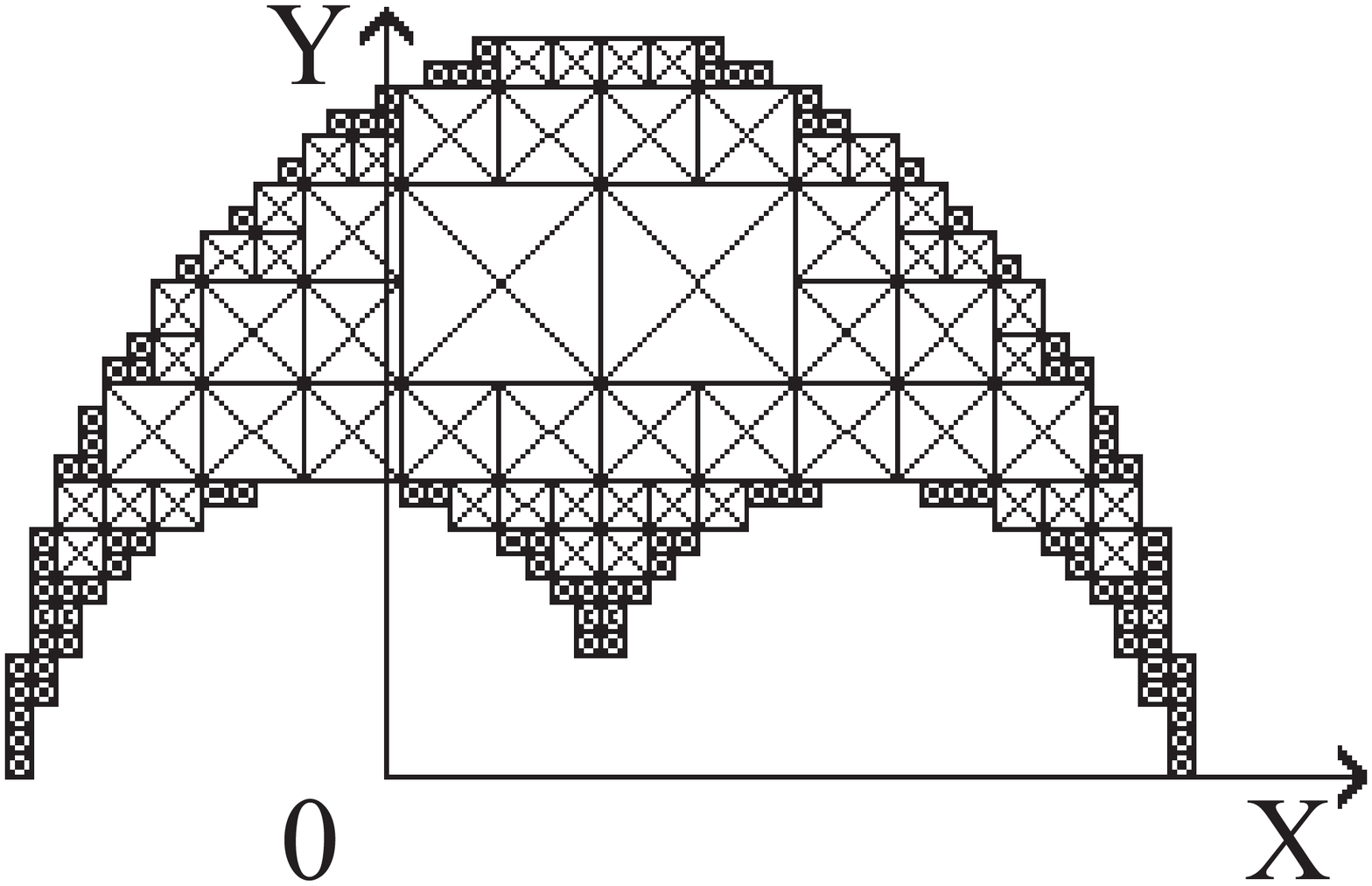}
           \caption{Joint configuration 2}
           \protect\label{figure:posture_2}
         \end{center}
       \end{minipage} &
       \begin{minipage}{36 mm}
         \begin{center}
           \includegraphics[width= 36mm,height= 22.8mm]{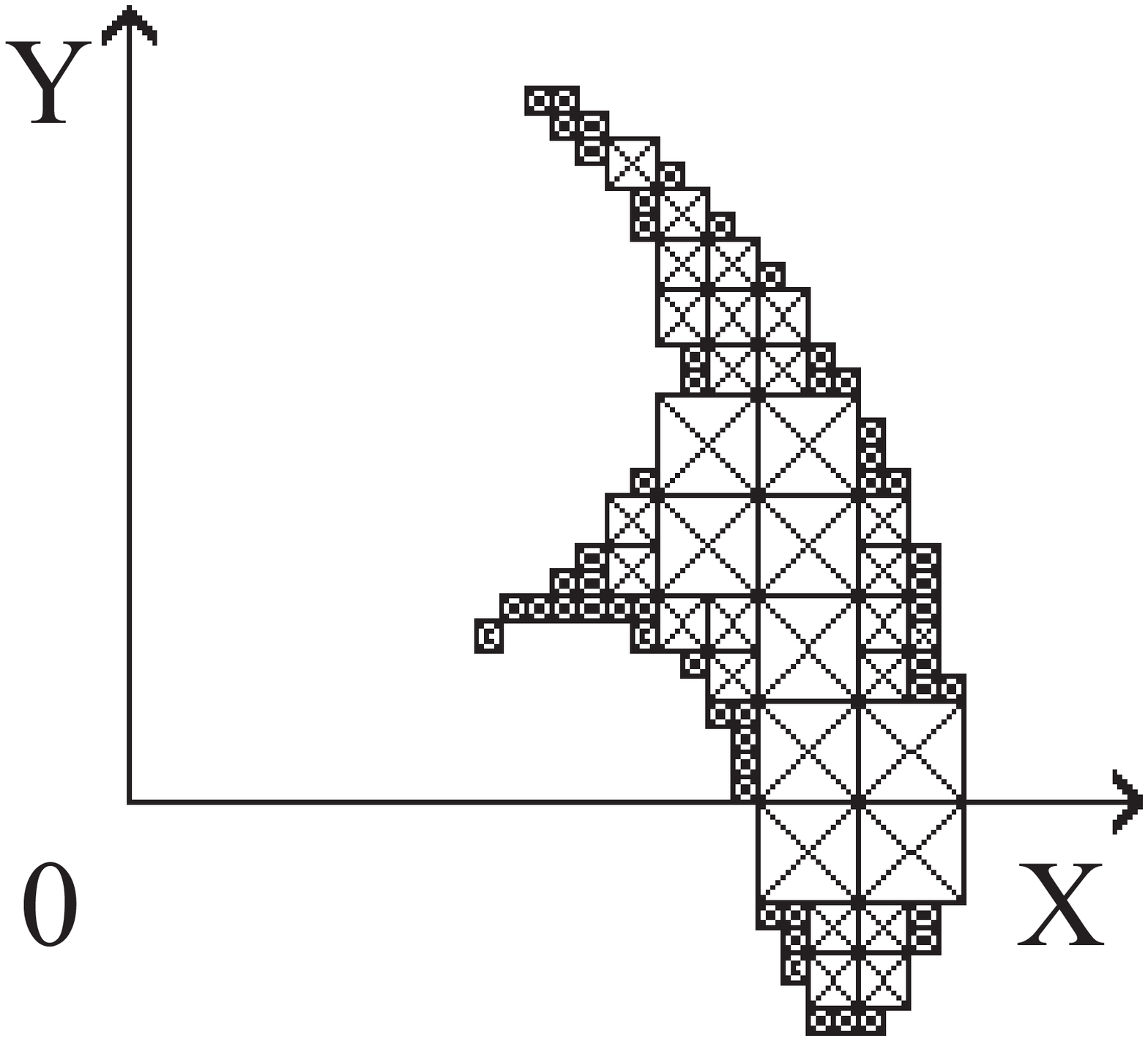}
           \caption{Joint configuration 3}
           \protect\label{figure:posture_3}
         \end{center}
       \end{minipage} &
       \begin{minipage}{36 mm}
         \begin{center}
           \includegraphics[width= 36mm,height= 22.8mm]{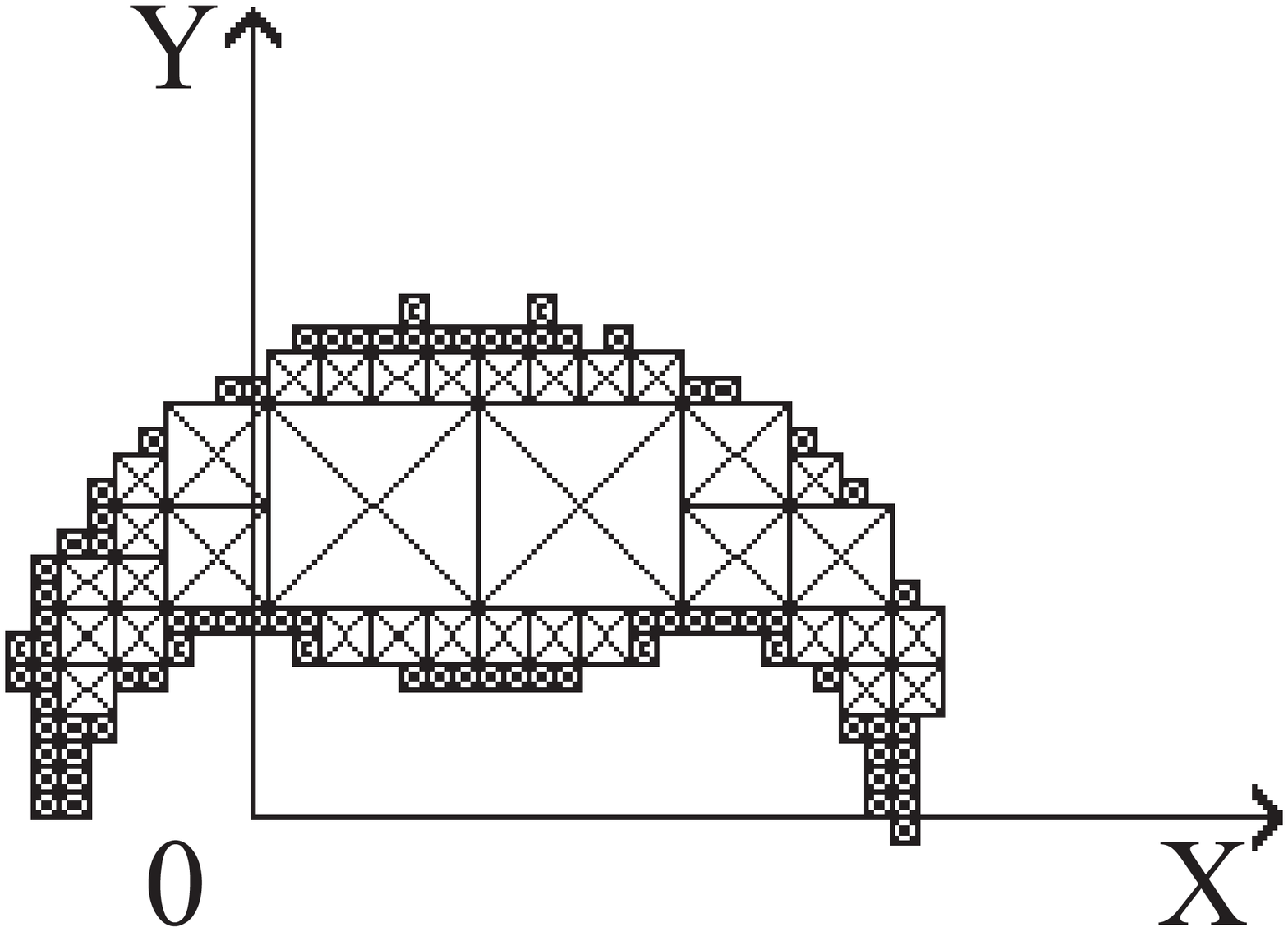}
           \caption{Joint configuration 4}
           \protect\label{figure:posture_4}
         \end{center}
       \end{minipage} \\
       \begin{minipage}{36 mm}
         \begin{center}
           \includegraphics[width= 36mm,height= 22.8mm]{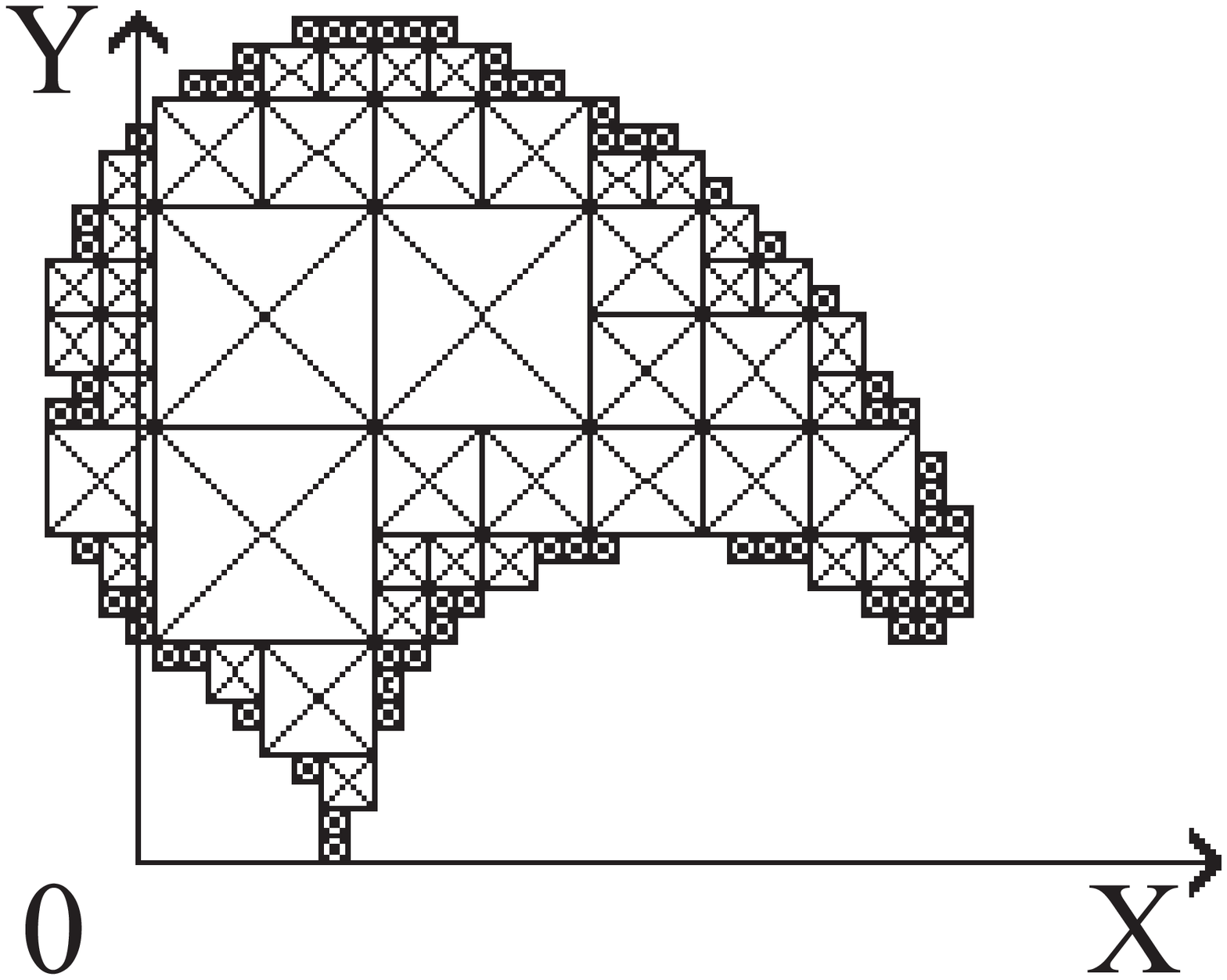}
           \caption{Joint configuration 5}
           \protect\label{figure:posture_5}
         \end{center}
       \end{minipage} &
       \begin{minipage}{36 mm}
         \begin{center}
           \includegraphics[width= 36mm,height= 22.8mm]{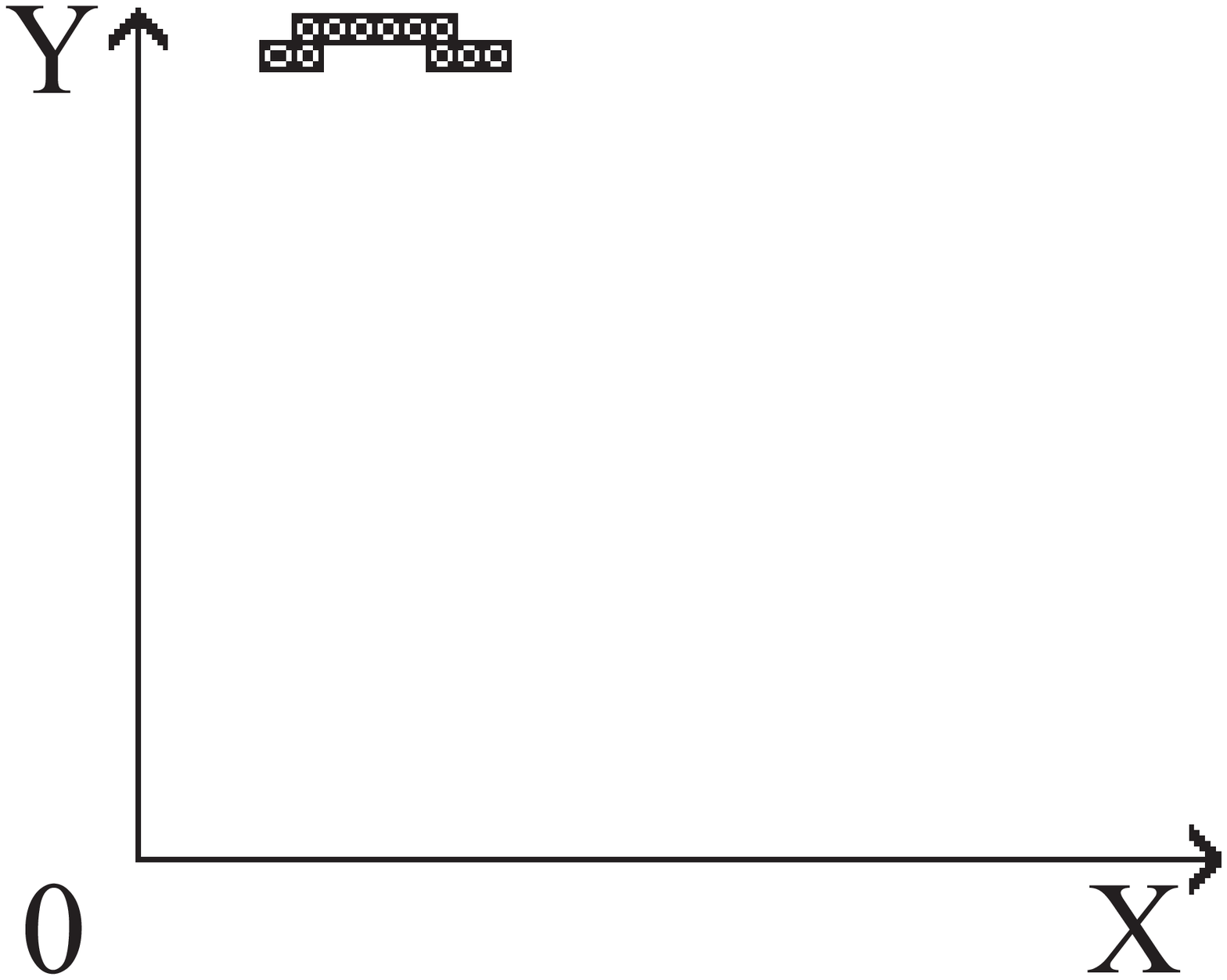}
           \caption{Joint configuration 6}
           \protect\label{figure:posture_6}
          \end{center}
       \end{minipage} &
       \begin{minipage}{36 mm}
         \begin{center}
           \includegraphics[width= 36mm,height= 22.8mm]{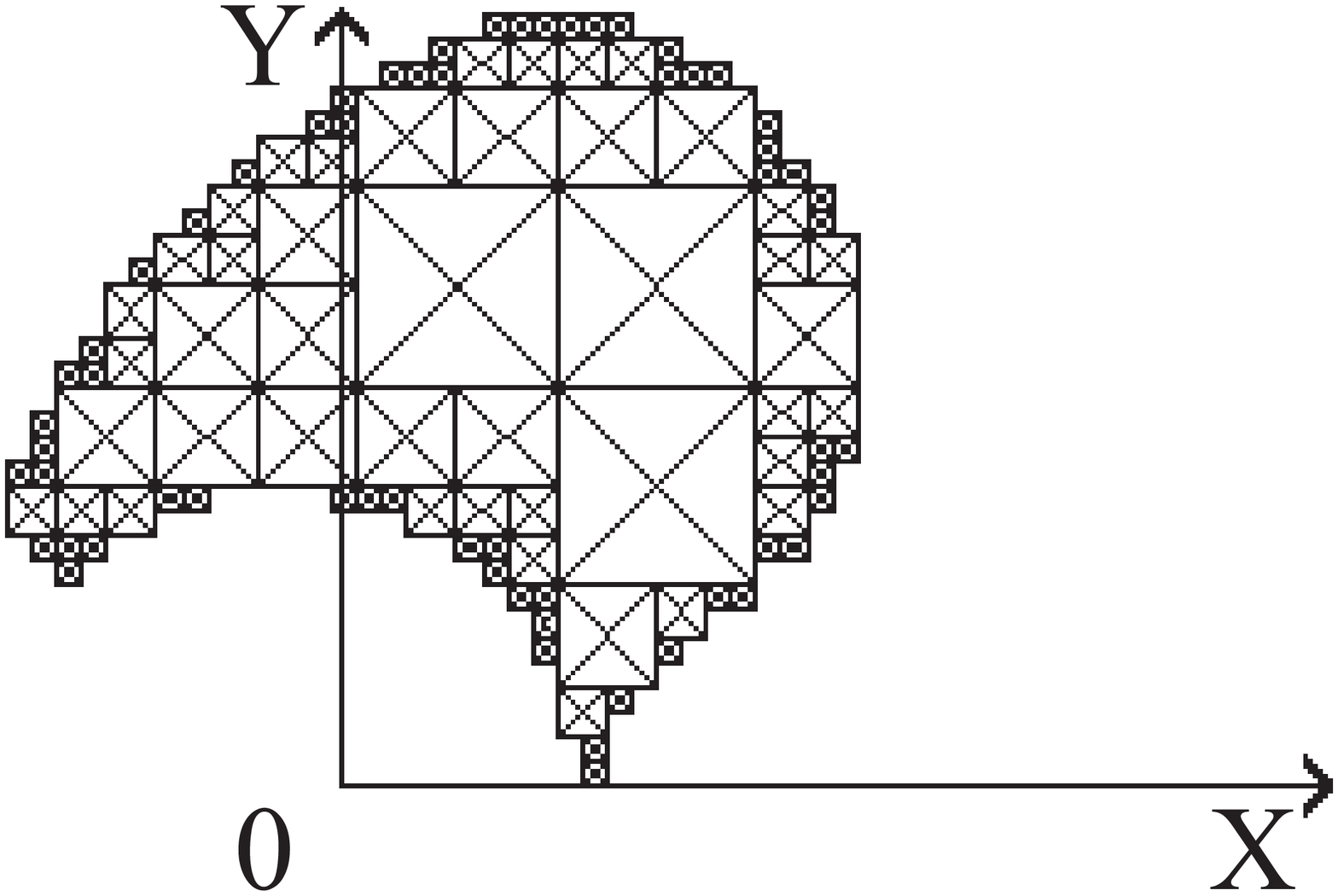}
           \caption{Joint configuration 7}
           \protect\label{figure:posture_7}
         \end{center}
       \end{minipage} &
       \begin{minipage}{36 mm}
         \begin{center}
           \includegraphics[width= 36mm,height= 22.8mm]{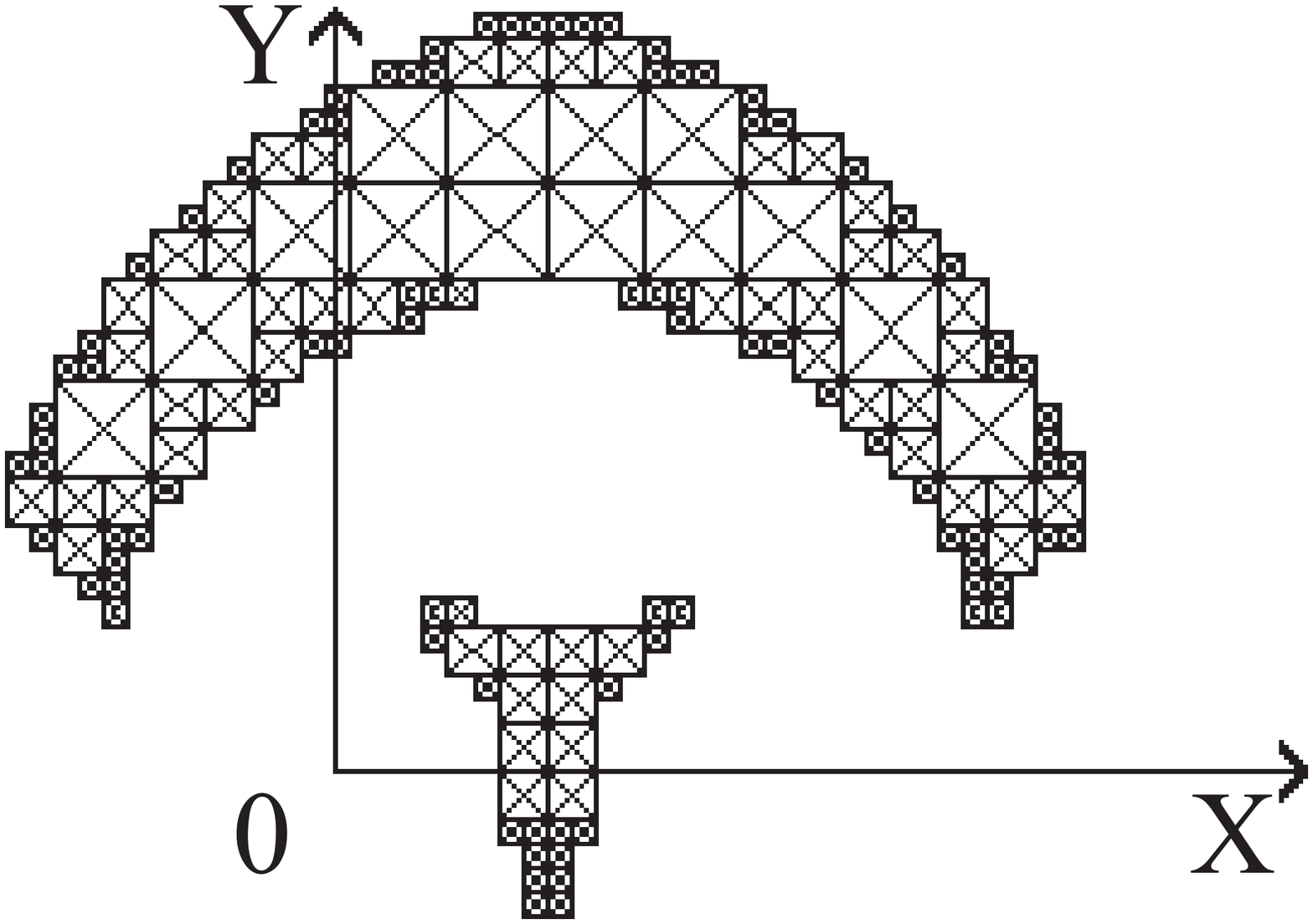}
           \caption{Joint configuration 8}
           \protect\label{figure:posture_8}
         \end{center}
       \end{minipage}
    \end{tabular}
\end{figure}

The T-connected regions are the projection onto the workspace of
the aspects. These regions are smaller than the N-connected
regions. According to the joint configuration of the manipulator,
i.e. according to the choice of the inverse kinematic solution, the
singularity-free regions differ. Figures \ref{figure:posture_1},
\ref{figure:posture_2}, \ref{figure:posture_3} and
\ref{figure:posture_4} depict the T-connected regions corresponding
to $det(\negr A) > 0$. Their union forms the first N-connected
region (Figure \ref{figure:First_N_connected_region}). Figures
\ref{figure:posture_5}, \ref{figure:posture_6},
\ref{figure:posture_7} and \ref{figure:posture_8} show the
T-connected regions such that $det(\negr A) < 0$ and their union is
the second N-connected region (Figure
\ref{figure:Second_N_connected_region}).
\section{Conclusions}
In this paper, the N-connected and the T-connected regions have
been defined to characterize the maximal regions of the workspace
where any point-to-point and continuous trajectories are feasible,
respectively.
\par
The manipulators considered in this study have multiple solutions
to their direct and inverse kinematics. The N-connected regions
were defined by first determining the maximum path-connected,
parallel singularity-free regions in the Cartesian product of the
workspace by the joint space. The projection of these regions onto
the workspace were shown to define the N-connected regions. The
T-connected regions were defined by the projection onto the
workspace of the aspects, i.e. the maximal serial and parallel
singularity-free domains in the Cartesian product of the workspace
by the joint space.
\par
The N-connectivity and the T-connected analysis of the workspace
are of high interest for the evaluation of manipulator global
performances as well as for off-line task programming.

\section{References}
~~~Chablat, D. and Wenger, Ph., ``Working Modes and Aspects in
Fully-Parallel Manipulator'', IEEE Int. Conf. On Robotics and
Automation, pp.~1964-1969, 1998.
\par
Chablat, D. and Wenger, Ph., ``Moveability and Collision Analysis
for Fully-Parallel Manipulators'', 12th CISM-IFTOMM Symposium,
RoManSy, 1998.
\par
Wenger, Ph., Chedmail, P., ``Ability of a Robot to Travel Through
its Free Workspace'', The Int. Jour. of Robotic Research, Vol.
10:3, June 1991.
\par
Gosselin, C.\ and Angeles, J., ``Singularity analysis of
closed-loop kinematic chains'', IEEE Trans. On Robotics And
Automation, Vol.~6, No.~3, June 1990.
\par
Nenchev, D.N., Bhattacharya, S., and Uchiyama, M.,``Dynamic
Analysis of Parallel Manipulators under the Singularity-Consistent
Parameterization'', Robotica, Vol.~15, pp.375-384,1997.
\par
Chablat, D., Wenger, Ph. , Angeles, J., ``The isoconditioning Loci
of A Class of Closed-Chain Manipulators'', IEEE Int. Conf. On
Robotics and Automation, pp. 1970-1975, May 1998.
\par
Chablat, D., ``Domaines d'unicit\'e et parcourabilit\'e pour les
manipulateurs pleinement parall\`eles'', PhD thesis, Nantes,
November 1998.
\end{document}